\documentclass[runningheads]{llncs}
\usepackage{amsmath}
\usepackage{amssymb}
\usepackage{xcolor}
\usepackage{placeins}
\usepackage{bbm}
\usepackage{multirow}
\usepackage{subcaption}
\usepackage{graphicx}
\usepackage{booktabs} 
\usepackage{longtable}
\allowdisplaybreaks
\usepackage[T1]{fontenc}
\usepackage{graphicx,verbatim}
\begin{document}

\title{FairQuant: Fairness-Aware Mixed-Precision Quantization  for Medical Image Classification}

%
\author{Thomas Woergaard\inst{1}
\and
Raghavendra Selvan\inst{1}
}
\authorrunning{Woergaard and Selvan}
%
\institute{Department of Computer Science, University of Copenhagen, Denmark
\email{thomaswoergaard@gmail.com, raghav@di.ku.dk}\\
\url{https://di.ku.dk/}}

\maketitle              

\vspace{-0.5cm}
\begin{abstract}
Compressing neural networks by quantizing model parameters offers useful trade-off between performance and efficiency. 
Methods like quantization-aware training and post-training quantization strive to maintain the downstream performance of compressed models compared to the full precision models. However, these techniques do not explicitly consider the impact on algorithmic fairness. In this work, we study fairness-aware mixed-precision quantization schemes for medical image classification under explicit bit budgets. We introduce FairQuant, a framework that combines group-aware importance analysis, budgeted mixed-precision allocation,
and a learnable Bit-Aware Quantization (BAQ) mode that jointly optimizes weights and per-unit bit allocations under bitrate and fairness regularization. We evaluate the method on Fitzpatrick17k and ISIC2019 across ResNet18/50, DeiT-Tiny, and TinyViT. Results show that FairQuant configurations with average precision near 4-6 bits recover much of the Uniform 8-bit accuracy while improving worst-group performance relative to Uniform 4- and 8-bit baselines, with comparable fairness metrics under shared budgets.
\footnote{Official source code at: \url{https://github.com/saintslab/FairQuant}.}
\end{abstract}
\vspace{-0.75cm}

\section{Introduction}
Deep neural networks are now routine in medical image analysis and reach strong performance in several tasks \cite{lecun2015deep,litjens2017survey}. 
In medical applications, subgroup reliability is a first-order requirement. 
Clinical datasets often under-represent certain groups, and performance gaps across minority groups can remain large even when the average accuracy is high~\cite{Groh2021Fitzpatrick17k,fitzscale2024}. 
A model can meet an average-accuracy target and still be unreliable for the worst-served group \cite{Hardt2016,barocas2019fairness}.

Resource costs and deployment constraints in clinical settings add pressure for efficient inference of deep learning models~\cite{selvan2022carbon}. Clinical decision support tools run under latency, memory, and energy budgets that favor low-precision arithmetic using model compression techniques such as quantization~\cite{selvan2023operating}. Quantization maps floating-point weights and activations to low-precision integers, reducing storage and enabling efficient integer operations~\cite{jacob2018quantization,gholami2021surveyquantizationmethodsefficient}. Mixed-precision quantization assigns different bit-widths to different parts of the network, keeping higher precision in sensitive units and compressing less sensitive units more aggressively~\cite{nagel2021white,Wenshoj2025CoDeQ}.

\begin{figure}[t]
    \centering
    \vspace{-0.25cm}
    \includegraphics[width=0.75\linewidth]{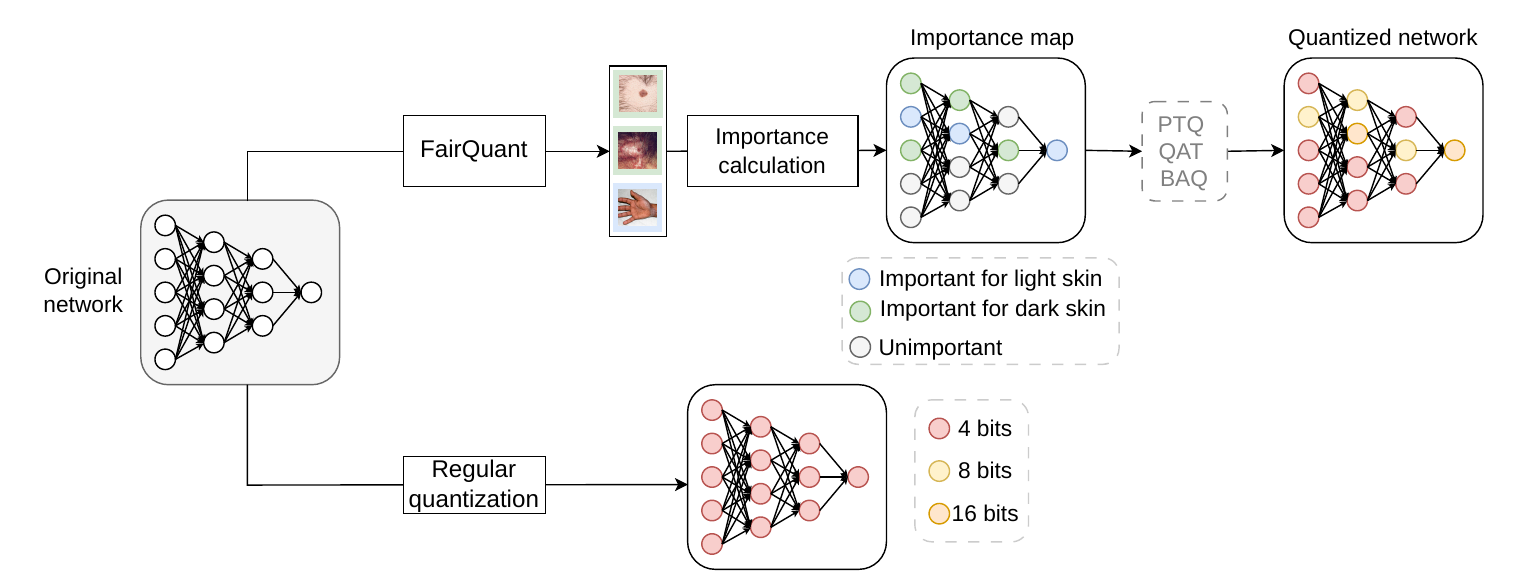}
    \vspace{-0.15cm}
    \caption{High-level overview of the proposed FairQuant framework (top pathway) illustrating the group importance calculation based on a calibration set, which is then used to allocate appropriate bit widths to the weights. Regular post-training quantization (bottom pathway).}
    \label{fig:fairquant}
    \vspace{-0.65cm}
\end{figure}

In this work, we study fairness-aware mixed-precision quantization with learnable bit allocation. We propose \emph{FairQuant}, a framework that couples group-aware importance analysis with a budgeted mixed-precision allocation rule. A short calibration stage collects group-conditioned sensitivity signals at the chosen quantization granularity, and a cross-group reducer converts these signals into a single importance map that reflects both overall sensitivity and group disparity. An allocation rule maps importance scores to discrete bit-widths under an explicit budget. On top of this allocation view, we introduce \emph{Bit-Aware Quantization} (BAQ), which treats per-unit bit-widths as learnable variables. BAQ optimizes weights and continuous bit proxies jointly under a bitrate regularizer and a fairness penalty. The learned proxies are discretized at the end of training to produce a mixed-precision model that meets the target budget. A high-level overview of the proposed framework is shown in Figure~\ref{fig:fairquant}.

We evaluate FairQuant on Fitzpatrick17k \cite{Groh2021Fitzpatrick17k} and ISIC 2019 \cite{ISIC2019,isicoverview2022}, spanning both convolutional (ResNet18/50 \cite{He2016resnet}) and transformer backbones (DeiT-Tiny~\cite{touvron2021training},  TinyViT \cite{wu2022tinyvitfastpretrainingdistillation}). 
Across both datasets and all four models, FairQuant yields a consistently better operating region at low average precision. At operating points around 4-6 average bits, FairQuant recovers much of the uniform 8-bit accuracy and improves worst-group accuracy relative to uniform 4-bit baselines, with improved group-conditional metrics under the same budget. The ablations vary the fairness-loss scale and FairQuant optimization settings to document stability and to expose how bitrate and fairness weights control the trade-off. \vspace{0.1cm}\\ 
{\bf Related Work.} 
Recent work has begun to connect fairness goals with model compression. Fairness-aware pruning and quantization have been done for dermatological diagnosis \cite{wu2022fairpruneachievingfairnesspruning,Guo2024}, and group-aware gradient signals have been used to guide pruning in face attribute classification in FairGRAPE~\cite{lin2022fairgrapefairnessawaregradientpruning}. Sensitivity-based mixed-precision methods are well established \cite{dong2019hawq,Dong2020}, yet most allocations are derived from global objectives that do not reflect subgroup risk. These approaches motivate compression schemes that incorporate group information. Two questions remain central for this study. Can a single approach improve subgroup outcomes under a strict bit budget without tailoring each model manually? Does the learned allocation remain stable across random seeds and training noise, so the reported trade-off reflects the method rather than a single run?

\section{Method}
\label{sec:method}

FairQuant allocates weight precision under an explicit budget using group conditioned sensitivity signals in two steps. 
A calibration pass computes an importance map per sensitive group at a chosen quantization granularity. A budgeted allocation rule maps the reduced importance map to a fixed mixed-precision pattern. BAQ then treats bit-widths as trainable variables and learns them jointly with weights under bitrate and fairness regularization as shown in Figure~\ref{fig:fairquant}. \vspace{0.1cm}\\ 
{\bf Setup and Quantization Operator.} 
The dataset is $\mathcal{D}=\{(x_i,y_i,g_i)\}_{i=1}^N$ with image $x_i$, label $y_i\in\{0,\ldots,C-1\}$, and group indicator $g_i\in\{0,..,G-1\}$. A model is $f(x;W)$ with weights $W=\{W_l\}_{l=1}^L$ and appropriate loss objectives $\ell(\cdot,\cdot)$.
We use symmetric uniform weight quantization \cite{jacob2018quantization}. A scope $\mathcal{S}$ is the unit that shares one common bit-width, induced by the granularity (per-tensor or per-channel). For bit-width $b\ge 2$, let $q_{\max}=2^{b-1}-1$. For scope weights $W_{\mathcal{S}}$ we compute the scale $s_{\mathcal{S}}=\lVert W_{\mathcal{S}}\rVert_\infty/q_{\max}$ and clamp $s_{\mathcal{S}}\leftarrow \max(s_{\mathcal{S}},\varepsilon)$, and define the quantize--dequantize map $Q_{b}(w;s_{\mathcal{S}})=s_{\mathcal{S}}\cdot \mathrm{clip}(\mathrm{round}(w/s_{\mathcal{S}}),-q_{\max},q_{\max})$. Rounding uses a straight-through estimator (STE)~\cite{bengio2013ste}, with the common surrogate $\partial Q_{b}(w)/\partial w \approx \mathbbm{1}(|w|\le q_{\max}s_{\mathcal{S}})$. \vspace{0.1cm}\\ 
{\bf Group-sensitive Importance Analysis.}
Calibration measures how strongly each group relies on each scope. The model is frozen and data is visited in mini-batches, with each backward pass using a group-restricted loss. For batch $j\in\{1,\ldots,B\}$ we define $S^{(j)}_g=\{i\in \text{batch }j \mid g_i=g\}$ and $n^{(j)}_g=|S^{(j)}_g|$, and for $n^{(j)}_g>0$ the group loss is $\mathcal{L}^{(j)}_g(W)=\frac{1}{n^{(j)}_g}\sum_{i\in S^{(j)}_g}\ell(f(x_i;W),y_i)$. FairQuant uses an importance score for measuring the importance of each scope, similar to FairGRAPE~\cite{lin2022fairgrapefairnessawaregradientpruning}, approximated using the first order Taylor expansion form in~\cite{molchanov2019importance}, given by 
\begin{equation}
I_{l,g}= \sum_{j=1}^{B}\big(\mathcal{L}^{(j)}_g(W) - \mathcal{L}^{(j)}_g(W|W_l=0)\big)^2 \approx \sum_{j=1}^{B}\big(\nabla_{W_l}\mathcal{L}^{(j)}_g(W)\odot W_l\big)^2,    
\end{equation}
yielding one tensor per layer and group. Importance tensors are then aggregated to the scope level. For convolutional weights of shape $(C_{\text{out}},C_{\text{in}},H,W)$
per-channel aggregation uses $\widetilde{I}_{l,g}[c]=\mathrm{mean}_{c',h,w}\,I_{l,g}[c,c',h,w]$ 
while per-tensor aggregation uses $\widetilde{I}_{l,g}=\mathrm{mean}\,I_{l,g}$. Group-conditioned maps are obtained
by computing $s_g=\sum_l\lVert \widetilde{I}_{l,g}\rVert_1$, normalizing $\bar{I}_{l,g}=\widetilde{I}_{l,g}/(\varepsilon+s_g)$, and taking $I_l=\max_g \bar{I}_{l,g}$, which yields a single importance map per layer at the chosen granularity~\cite{lin2022fairgrapefairnessawaregradientpruning}.\vspace{0.1cm}\\ 
{\bf Budgeted Mixed-Precision Allocation.}
Budgeted allocation maps importance values to discrete bits under a global bit budget. We concatenate layer-level importance maps as $\mathbf{v}=[I_1,\ldots,I_L]\in\mathbb{R}^M$ with $M=\sum_{l=1}^L M_l$, where $M_l$ is the number of scopes in layer $l$. We choose an admissible bit budget palette $\mathbf{b}=\{b_1<b_2<\cdots<b_K\}$ and target proportions $\boldsymbol{\pi}=(\pi_1,\ldots,\pi_K)$ with $\sum_{k=1}^{K}\pi_k=1$, define cumulative levels $c_k=\sum_{j=1}^{k}\pi_j$ for $k=1,\ldots,K-1$, and set thresholds $t_k=\mathcal{Q}_{c_k}(\mathbf{v})$ where $\mathcal{Q}_\tau(\mathbf{v})$ is the empirical $\tau$-quantile. These thresholds induce a tiering map $\phi(v_i)$ that assigns $b_1$ if $v_i\le t_1$, assigns $b_k$ if $t_{k-1}<v_i\le t_k$ for $k=2,\ldots,K-1$, and assigns $b_K$ if $v_i>t_{K-1}$. Applying $\phi$ yields an assignment vector $\mathbf{a}\in\{b_1,\ldots,b_K\}^M$, reshaped back to per-layer scopes $\{a_{\mathcal{S}}\}$; this fixed pattern can be used as an optional warm start for BAQ.\vspace{0.1cm}\\ 
{\bf Bit-Aware Quantization (BAQ).} 
BAQ turns scope bit-widths into trainable parameters, based on the formulation in~\cite{Wenshoj2025CoDeQ}. Each scope $\mathcal{S}$ carries a real logit $b^{\mathrm{logit}}_{\mathcal{S}}\in\mathbb{R}$ that maps to a continuous bit proxy
\begin{align}
b^{\mathrm{cont}}_{\mathcal{S}}
=
\tanh\big(|b^{\mathrm{logit}}_{\mathcal{S}}|\big)(b_{\max}-b_{\min})+b_{\min},
\qquad
b_{\mathcal{S}}=
\mathrm{round}\big(b^{\mathrm{cont}}_{\mathcal{S}}\big),
\label{eq:baq_bits}
\end{align}
with $2\le b_{\min}<b_{\max}$. Gradients for the rounding operation is approximated using STE~\cite{bengio2013ste}. In each forward pass, scope $\mathcal{S}$ is quantized with $Q_{b}(W;s_{\mathcal{S}})$ using $b_{\mathcal{S}}$, with $s_{\mathcal{S}}$ recomputed from the current floating weights. Backpropagation updates both floating weights and bit logits; gradients reach the logits through the quantized loss and through the bitrate regularizer below. At the end of training, $b_{\mathcal{S}}$ defines the inference-time mixed-precision model.

Training minimizes three components related to the task, fairness, and bitrate. For a mini-batch of size $B$,
\begin{align}
\mathcal{L}_{\text{task}}
&=\frac{1}{B}\sum_{k=1}^{B}\ell\big(f(x_i;Q(W)),y_i\big),\label{eq:task_loss}\\
\mathcal{L}_g
&=\frac{1}{|S_g|}\sum_{i\in S_g}\ell\big(f(x_i;Q(W)),y_i\big), \qquad S_g = \{ i : g_i = g \}\label{eq:group_batch_loss}\\
\mathcal{L}_{\text{fair}}
&=\max_g \mathcal{L}_g-\min_g \mathcal{L}_g.\label{eq:fair_loss}
\end{align}
The total loss starts from
 $\mathcal{L}_{\text{base}} = \mathcal{L}_{\text{task}} +
\lambda_{\text{fair}}\mathcal{L}_{\text{fair}}$,
where $\lambda_{\text{fair}}\ge 0$ is shared with the non-learnable quantization modes. BAQ adds an $L2$ penalty on the logits that control the bits. Let $\{b^{\mathrm{logit}}_{\mathcal{S}}\}$ collect all BAQ logits in the network. The regularizer is given as,
$\mathcal{L}_{\text{baq},b}
=
\sum_{\mathcal{S}}
\left\| b^{\mathrm{logit}}_{\mathcal{S}} \right\|_2^2$,
weighted by a chosen coefficient $\lambda_{\text{baq},b}\ge 0$. The final loss is
$\mathcal{L}
=
\mathcal{L}_{\text{base}}
+
\lambda_{\text{baq},b}\mathcal{L}_{\text{baq},b}$.
where $\lambda_{\text{fair}}$ controls the pressure toward smaller group loss spread and $\lambda_{\text{baq},b}$ controls the precision regime by shrinking $b^{\mathrm{logit}}_{\mathcal{S}}$ toward zero, which pushes $b^{\mathrm{cont}}_{\mathcal{S}}$ toward $b_{\min}$.

Initialization uses either a high-precision start (large logits so $b^{\mathrm{cont}}_{\mathcal{S}}\approx b_{\max}$) or a warm start from the static allocation. For warm start, the fixed assignment $a_{\mathcal{S}}$ is converted into logits that reproduce $a_{\mathcal{S}}$ at the first BAQ forward pass \cite{Wenshoj2025CoDeQ}. After training, the rounded bits $b_{\mathcal{S}}$ are used for inference.

\vspace{-0.25cm}
\section{Experiments}
\label{sec:experiments}
\vspace{-0.15cm}
{\bf Datasets and Sensitive Attributes.}
The study uses two dermatology benchmarks with annotated sensitive attributes. Fitzpatrick17k contains 16,577 clinical photographs with 114 diagnostic categories and Fitzpatrick skin type labels~\cite{Groh2021Fitzpatrick17k,fitzscale2024}. The sensitive attribute is Fitzpatrick type, and we also uses a binary grouping with types I-III (lighter) versus types IV-VI (darker), following prior dermatology fairness work \cite{Guo2024}. 

ISIC 2019 contains 25,331 dermoscopic images across nine diagnostic categories \cite{ISIC2019,isicoverview2022}. 
The sensitive attribute is patient sex from the metadata file, yielding three groups: female, male, and unknown.
Training-time preprocessing applies resize, random horizontal and vertical flips, random rotation, and AutoAugment policy, followed by normalization with ImageNet statistics~\cite{paszke2019pytorchimperativestylehighperformance}. \vspace{0.1cm}\\
{\bf Architectures and Training Protocol.}
We evaluate two convolutional backbones (ResNet18/50) \cite{He2016resnet} and two compact vision transformers (DeiT-Tiny, TinyViT)\cite{touvron2021training,wu2022tinyvitfastpretrainingdistillation}. Each backbone is initialized from ImageNet-pretrained weights \cite{deng2009imagenet,torchvision2016}. The classifier head is replaced to match the dataset label space, then the full-precision reference model is obtained by fine-tuning on the training split.

Optimization uses AdamW \cite{loshchilov2019decoupledweightdecayregularization} with cross-entropy loss, batch size 128, weight decay 0.01, and a fixed learning rate of $10^{-4}$ up to epoch 160, followed by a $\times0.1$ decay for the remaining epochs. ResNet models are pretrained for 200 epochs; DeiT-Tiny and TinyViT are pretrained for 20 epochs. 
\vspace{0.1cm}\\
{\bf Quantization Regimes and Evaluation Grid.}
All quantization runs start from the full-precision checkpoint for the same dataset and backbone. We quantize only the weights in convolutional and linear layers. 
We use symmetric uniform quantization with gradients from STE \cite{jacob2018quantization,bengio2013ste} at per-channel granularity.

We focus on the low-precision regime where uniform quantization baselines can lose accuracy and reduce the worst-group performance. The main comparison set is FP32, Uniform-8 and Uniform-4 as the post-training quantization (PTQ) baselines, fixed mixed-precision QAT based on the FairQuant importance signal, and FairQuant with BAQ for learned mixed precision. PTQ baselines evaluate the quantized checkpoint without additional training. QAT and FairQuant fine-tune for 10 epochs using the same optimizer, batch size, data pipeline, and learning-rate policy as the full-precision run.

FairQuant importance signals are computed on a calibration set drawn from the training split. We use 50 mini-batches with the same batch size and augmentations as training. Gradients are accumulated per sensitive group to form group-conditioned importance maps. These maps are reduced across groups and converted into a bit assignment under the chosen bit palette and budget. The static assignment initializes fixed mixed-precision QAT and can warm-start BAQ. BAQ uses a fixed bit interval $[b_{\min},b_{\max}]$ and learns scope-level bits within this range.
\vspace{0.1cm}\\
{\bf Metrics and Ablations.}
We report average accuracy (AvgAcc), worst-group accuracy (WorstAcc), equalised opportunity gap (EOpp0) and equalized odds gap (EOdd)~\cite{Hardt2016}.
Computational efficiency is quantified using effective bits per model parameter and giga bit-operations (GBOPs) from Wang et al. \cite{wang2021differentiablejointpruningquantization}.

Main comparisons use three seeds per dataset, model, and method, and we report means and standard deviation for accuracy measures. Ablations use five seeds per setting and report means with 95\% confidence intervals. We run three targeted ablations. The first sweeps the BAQ bitrate regularizer $\lambda_{\text{baq},b}$ to assess stability and to trace the accuracy and fairness trade-off, the second test  fairness-loss scale $\lambda_{\text{fair}}$, and the third sweeps the stability by varying the learning rate used for the BAQ
 on ResNet18 for both datasets shown in Fig.~\ref{fig:ablation}. 

\vspace{-0.15cm}
\section{Results}
\label{sec:results}
\vspace{-0.15cm}

\begin{figure}[t]
\vspace{-0.25cm}
    \centering
    \begin{subfigure}{0.45\textwidth}
            \includegraphics[width=\linewidth]{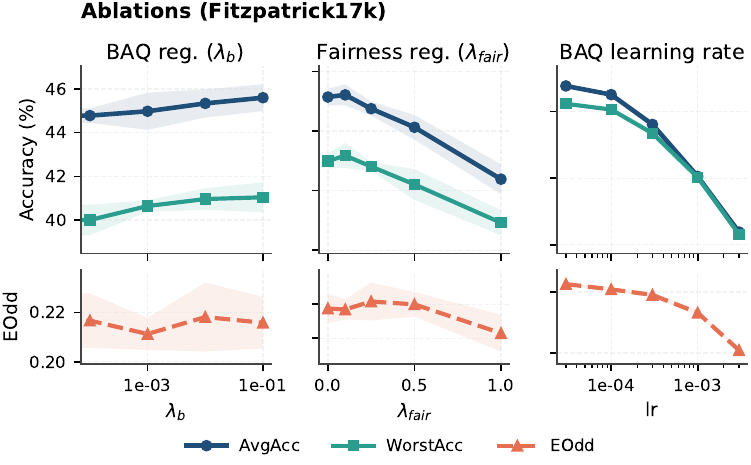}
            \caption{}
    \end{subfigure}
    \begin{subfigure}{0.45\textwidth}
            \includegraphics[width=\linewidth]{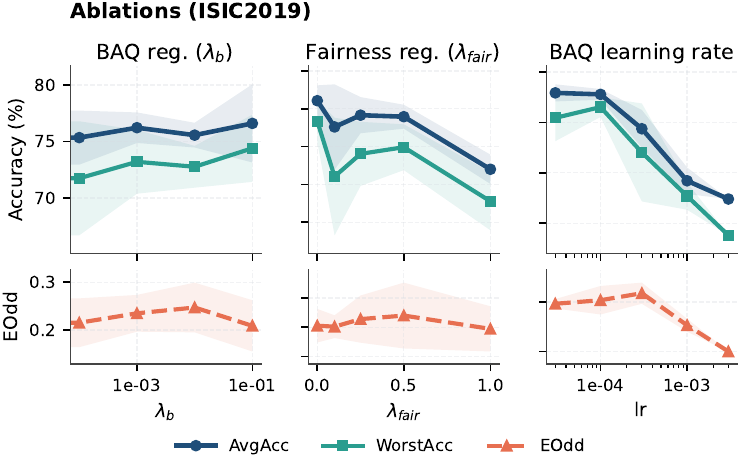}
            \caption{}
    \end{subfigure}
    \vspace{-0.25cm}
    \caption{Ablations and stability for BAQ. Left: sweep of the bitrate regularizer \(\lambda_{\mathrm{baq},b}\) showing test AvgAcc and EOpp on Fitzpatrick17k and ISIC2019. Middle: sweep of the fairness-loss scale \(\lambda_{\mathrm{fair}}\) on the ResNet-18 backbone. Right: learning-rate sweep for BAQ bit proxies. All panels use a fixed BAQ bit interval \([b_{\min},b_{\max}]~=[4,16]\). Shaded regions indicate variability across repeated runs.}
    \label{fig:ablation}
    \vspace{-0.25cm}
\end{figure}

\begin{table}[t]
\vspace{-0.35cm}
\tiny
\centering
\caption{Main results across two datasets and different models. All QAT experiments use the following bit allocation \{20\%: 2, 40\%: 4, 40\%: 8\}, and BAQ is reported for range 4-8. Significant results for average bits, average accuracy, and worst group accuracy compared to the baseline methods are shown in bold.}
\vspace{0.1cm}
\label{tab:combined_results}
\resizebox{\textwidth}{!}{%
\begin{tabular}{cccrclllcc}
\toprule
{\bf Dataset} & {\bf Model} & {\bf Method} & {\bf AvgBits}$\downarrow$ & {\bf GBOPs}$\downarrow$ & {\bf AvgAcc}$\uparrow$ & {\bf WorstAcc}$\uparrow$ & {\bf Gap}$\downarrow$ & {\bf EOpp0}$\downarrow$ & {\bf EOdd}$\downarrow$ \\
\midrule
\multirow{15}{*}{\bf FitzP.} & \multirow{5}{*}{DeiT-Tiny} & FP32  & 32.00 & 0.07 & 53.00 & 47.00 & 12.10 & 0.003 & 0.225 \\
 & & U8  & 8.00 & 0.02 & 53.00 & 47.00 & 12.00 & 0.003 & 0.224 \\
 & & U4  & 4.00 & 0.01 & 45.50 & 40.20 & 10.60 & 0.003 & 0.222 \\\cmidrule{3-10}
& & Lin et al.\cite{lin2022fairgrapefairnessawaregradientpruning}  & 4.80 & 0.01 & 32.60 & 28.70 & 7.80 & 0.004 & 0.141 \\
& & FQ-QAT & 5.03 & 0.01 & {\bf 54.10}$_{\pm0.46}$ & {\bf 47.90}$_{\pm0.56}$ & 12.30$_{\pm 0.72}$& 0.003  & 0.229  \\
& & FQ-BAQ  & {\bf 4.30} & 0.01 & 51.83$_{\pm 0.64}$ & 46.00$_{\pm0.10}$ & 11.67$_{\pm 1.31}$& 0.003 & 0.220 \\
\cmidrule{2-10}
& \multirow{5}{*}{TinyViT} & FP32  & 32.00 & 0.75 & 56.30 & 50.00 & 12.60 & 0.003 & 0.227 \\
 & & U8  & 8.00 & 0.19 & 55.50 & 49.50 & 12.00 & 0.003 & 0.234 \\
 &  & U4  & 4.00 & 0.09 & 3.00 & 2.00 & 1.90 & 0.004 & 0.015 \\\cmidrule{3-10}
& & Lin et al.\cite{lin2022fairgrapefairnessawaregradientpruning}  & 4.80 & 0.11 & 13.40 & 12.00 & 2.20 & 0.003 & 0.041 \\
& & FQ-QAT  & 4.43 & 0.18  & 49.27$_{\pm0.40}$ & 44.53$_{\pm0.47 }$& 9.53$_{\pm 0.21} $& 0.003   & 0.201  \\
& & FQ-BAQ  & {\bf 4.12} & 0.13 & {\bf 53.60}$_{\pm0.26}$ & {\bf 48.17}$_{\pm0.12}$ & 10.93$_{\pm 0.67}$& 0.003  & 0.239 \\
\cmidrule{2-10}
& \multirow{5}{*}{ResNet18} & FP32  & 32.00 & 3.63 & 50.60 & 44.10 & 12.90 & 0.003 & 0.228 \\
& & U8  & 8.00 & 0.91 & 50.50 & 44.00 & 13.00 & 0.003 & 0.232 \\
& & U4  & 4.00 & 0.45 & 23.40 & 19.00 & 8.90 & 0.004 & 0.162 \\\cmidrule{3-10}
& & Lin et al.\cite{lin2022fairgrapefairnessawaregradientpruning}  & 4.80 & 0.54 & 44.70 & 40.70 & 7.00 & 0.004 & 0.182 \\
& & Guo et al.\cite{Guo2024}  & 4.92 & 0.56 & 40.80 & 36.60 & 14.90 & 0.003 & 0.216 \\
& & FQ-QAT & {\bf 3.77} & 0.70 & 40.73$_{\pm0.25}$ & 36.07$_{\pm0.45}$ & 9.37$_{\pm 0.40}$ & 0.004  & 0.180  \\
& & FQ-BAQ  & 4.07 & 0.57 & {\bf 45.33}$_{\pm0.93} $& {\bf 41.53}$_{\pm0.25 }$& 7.60$_{\pm 1.73}$ & 0.003 & 0.214  \\
\midrule
\multirow{15}{*}{\bf ISIC} & \multirow{5}{*}{DeiT-Tiny} & FP32  & 32.00 & 0.07 & 81.30 & 81.00 & 0.80 & 0.037 & 0.198 \\
 &  & U8  & 8.00 & 0.02 & 81.30 & 81.00 & 0.70 & 0.036 & 0.199 \\
 & & U4  & 4.00 & 0.01 & 81.40 & 79.10 & 5.70 & 0.035 & 0.238 \\
 \cmidrule{3-10}
&  & Lin et al.\cite{lin2022fairgrapefairnessawaregradientpruning}  & 4.80 & 0.01 & 74.50 & 69.60 & 8.60 & 0.061 & 0.236 \\
&  & FQ-QAT & 5.04 & 0.01 & {\bf 83.80}$_{\pm0.26}$ & {\bf 82.87}$_{\pm0.12}$& 2.37$_{\pm 0.64}$ & 0.033  & 0.160 \\
&  & FQ-BAQ  & {\bf 4.35} & 0.01 & 81.13$_{\pm1.42}$ & 79.30$_{\pm1.93} $& 4.20$_{\pm 2.00}$ & 0.031 & 0.198 \\
\cmidrule{2-10}
& \multirow{5}{*}{TinyViT} & FP32  & 32.00 & 0.75 & 84.80 & 82.90 & 4.40 & 0.029 & 0.147 \\
& & U8  & 8.00 & 0.19 & 84.60 & 82.50 & 4.80 & 0.030 & 0.163 \\ 
 & & U4  & 4.00 & 0.09 & 53.30 & 47.60 & 9.10 & 0.014 & 0.046 \\
 \cmidrule{3-10}
& & Lin et al.\cite{lin2022fairgrapefairnessawaregradientpruning}  & 4.80 & 0.11 & 66.10 & 64.60 & 3.90 & 0.041 & 0.132 \\
&  & FQ-QAT & 4.38 & 0.18 & {\bf 83.40}$_{\pm 0.20}$ & {\bf 82.47}$_{\pm0.25}$ & 1.90$_{\pm0.72}$ & 0.035  & 0.162  \\
&  & FQ-BAQ  & {\bf 4.13} & 0.14 & 82.73$_{\pm0.97}$ & 81.13$_{\pm0.61}$ & 3.07$_{\pm1.16}$ & 0.035 & 0.184  \\
\cmidrule{2-10}
& \multirow{5}{*}{ResNet50} & FP32  & 32.00 & 8.17 & 76.80 & 75.80 & 2.70 & 0.026 & 0.161 \\
& & U8  & 8.00 & 2.04 & 76.40 & 75.90 & 1.30 & 0.030 & 0.174 \\ 
 & & U4  & 4.00 & 1.02 & 57.80 & 51.90 & 13.90 & 0.021 & 0.145 \\
 \cmidrule{3-10}
& & Lin et al.\cite{lin2022fairgrapefairnessawaregradientpruning}  & 4.80 & 1.23 & {\bf 86.20} & 79.70 & 10.50 & 0.034 & 0.178 \\
& & FQ-QAT & 4.39 & 1.48 & 84.87$_{\pm0.21}$ & {\bf 82.67}$_{\pm0.29}$ & 5.07$_{\pm0.93}$ & 0.032 & 0.171  \\
& & FQ-BAQ  & {\bf 4.11} & 1.26 & 82.17$_{\pm1.12}$ & 80.83$_{\pm2.04}$ & 2.63$_{\pm 1.69}$ & 0.033 & 0.220  \\
\bottomrule
\end{tabular}%
}
\vspace{-0.5cm}
\end{table}
In Table~\ref{tab:combined_results} we report the performance of FairQuant with QAT (FQ-QAT) and BAQ (FQ-BAQ) on the two datasets and different architectures, 
comparing them to FP32 and Uniform 8-bit and Uniform 4-bit models along with FairGRAPE~\cite{lin2022fairgrapefairnessawaregradientpruning}. Performance of Uniform 8-bit quantization is close to FP32 across datasets and architectures, whereas Uniform 4-bit can be unreliable.
\vspace{0.1cm}\\
{\em Performance on Fitzpatrik17k.} ResNet18 with uniform 4-bit drops AvgAcc from 50.6 to 23.4, with WorstAcc 19.0. On the same setting, FQ-BAQ at 4.07 average bits recovers AvgAcc 45.33 and WorstAcc 41.53. For TinyViT on Fitzpatrick17k, Uniform 4-bit collapses to AvgAcc 3.0 and WorstAcc 2.0, while FQ-BAQ at 4.12 average bits reaches AvgAcc 53.6 and WorstAcc 48.17, approaching the FP32/U8 regime. For DeiT-Tiny on Fitzpatrick17k, Uniform 4-bit reaches AvgAcc 45.5 and WorstAcc 40.2, while FQ-BAQ at 4.30 average bits reaches AvgAcc 51.83  and WorstAcc 46.00.
\vspace{0.1cm}\\
{\em Performance on ISIC2019.} Similar trend is observed on ISIC2019. Uniform 4-bit can either collapse (e.g. TinyViT: AvgAcc 53.3, WorstAcc 47.6; ResNet50: AvgAcc 57.8, WorstAcc 51.9) or worsen subgroup reliability even when AvgAcc is preserved (Deit-Tiny: AvgAcc 81.4 but WorstAcc 79.1 and Gap 5.7). In contrast, FQ-BAQ at roughly 4.1-4.4 average bits consistently recovers performance in these low-bit regimes (e.g. TinyViT: AvgAcc 82.73, WorstAcc 81.13; ResNet50: AvgAcc 82.17, WorstAcc 80.83), while FQ-QAT achieves the highest accuracies at slightly higher average precision (e.g. DeiT-Tiny: AvgAcc 83.80, WorstAcc 82.87; Table~\ref{tab:combined_results}). These results place FairQuant on a more favourable frontier in the low-bit regime, where Uniform 4-bit quantization often trades off a large amount of AvgAcc and WorstAcc for a relatively small reduction in average precision. Note that on ISIC, the pretrained ResNet50 baseline and Uniform quantization has lower average accuracy compared to the other quantization methods, which we expect is due to the additional training performed during QAT and BAQ; similar increase is also reported in FairQuantize~\cite{Guo2024}.
\vspace{0.1cm}\\
{\bf Ablations and Stability.}
We analyze robustness of BAQ with respect to its main control parameters and training dynamics. The ablations are designed to 
justify the default hyperparameters used in the main grid. All ablations have been done with a ResNet-18.
\vspace{0.1cm}\\
{\em BAQ bitrate regularization.}
Fig.~\ref{fig:ablation}-a and Fig.~\ref{fig:ablation}-b (left) sweeps the BAQ bitrate regularizer weight \(\lambda_{\text{baq},b}\). Across both datasets, AvgAcc and WorstAcc vary only modestly over the tested range, and the shaded bands stay tight, indicating low sensitivity across seeds. EOdd remains within a narrow band across the sweep, with no signs of instability. These trends support choosing a default value from the broad flat region, and we use \(\lambda_{\text{baq},b} = 0.01\) for all main experiments. 
\vspace{0.1cm}\\
{\em Fairness regularization.}
Fig.~\ref{fig:ablation}-a and Fig.~\ref{fig:ablation}-b (middle) sweeps the fairness regularization \(\lambda_{\mathrm{fair}}\). As \(\lambda_{\mathrm{fair}}\) increases, EOdd generally improves, with the clearest gains at the high end of the sweep, paired with a noticeable drop in both AvgAcc and WorstAcc.  Part of this apparent improvement in EOdd at large \(\lambda_{\mathrm{fair}}\) may reflect that the classifier is becoming less accurate overall, so errors increase across groups and the performance gap narrows, yielding a lower disparity rather than a strictly better operating point.  For intermediate \(\lambda_{\mathrm{fair}}\), the figure suggests small fairness gains with limited accuracy loss, and we pick \(\lambda_{\mathrm{fair}}=0.5\) in this moderate regime for the default setting.
\vspace{0.1cm}\\
{\em BAQ optimizer stability.}
Fig.~\ref{fig:ablation}-a and Fig.~\ref{fig:ablation}-b (right) varies the BAQ learning rate used for the bit proxies. Performance is stable in the low learning rate regime, then degrades as the learning rate increases, with both AvgAcc and WorstAcc dropping sharply at the largest values and EOdd shifts at the same time, reflecting a different operating point rather than noisy runs. The sweep shows that BAQ does not require fine tuning within a narrow learning rate window, as long as learning rates remain in the stable low range.

\section{Discussion and Conclusions}
\label{sec:discussion}

The results show that precision allocation is not a purely technical detail once group information enters the objective. In the dermatology setting, low-precision baselines can fail in a way that is invisible from average accuracy alone. Uniform 8-bit tracks FP32 closely, yet Uniform 4-bit can collapse for certain backbones, with large drops in worst-group accuracy on both datasets. FairQuant shifts this operating region. 

The group-conditioned importance signal and the balanced reducer provide a stable starting point, so static mixed-precision patterns already improve over uniform baselines in several settings. BAQ then turns allocation into an optimization problem such that bit-widths become parameters that move during training under explicit bitrate control and a fairness penalty. The stability ablation over the BAQ bitrate regularizer supports a practical takeaway in the way that the method does not require narrow hyperparameter tuning.
\vspace{0.1cm}\\
{\bf Limitations.}
The evaluation is limited to two dermatology datasets and two sensitive attributes (skin type and sex), where group labels can be noisy or incomplete \cite{Groh2021Fitzpatrick17k}. The fairness term uses a batch-level proxy for group disparities, and the reported fairness metric focuses on EOpp and EOdd; other definitions can change which operating point is preferred \cite{Hardt2016,barocas2019fairness}. A limitation is that we only ran a coarse sweep over $\lambda_{\mathrm{fair}}$; a denser and better targeted sweep may reveal a clearer picture of its effect and identify a stronger operating point.
\vspace{0.1cm}\\
{\bf Conclusions.} \label{sec:conclusion}
This paper presents FairQuant, a fairness-aware mixed-precision framework that couples group-conditioned importance analysis with an explicit budgeted allocation rule, and BAQ, a learnable bit-width training objective with bitrate control and a fairness penalty. On Fitzpatrick17k and ISIC2019, across ResNet18/50 and DeiT-Tiny/TinyViT, the method improves the low-precision operating region where uniform 4-bit quantization can degrade worst-group performance. The resulting frontier provides a practical path to compressed models with stronger subgroup reliability under a fixed precision budget.

\clearpage
\bibliographystyle{splncs04}
\bibliography{references}

\end{document}